\def\year{2017}\relax
\documentclass[letterpaper]{article} 
\usepackage{aaai17}  
\usepackage{times}  
\usepackage{helvet}  
\usepackage{courier}  
\usepackage{url}  
\usepackage{graphicx}  

\usepackage{xcolor} 
\usepackage{relsize}

\newcommand{\FOtwo}{\mathit{FO}^2}
\newcommand{\StwoFOtwo}{\mathit{S}^2\mathit{FO}^2}
\newcommand{\StwoRU}{\mathit{S}^2\mathit{RU}}

\newcommand{\pop}[1]{\Delta_{#1}}
\newcommand{\lv}[1]{#1}
\newcommand{\lvs}[1]{\mathbf{#1}}
\newcommand{\ind}[1]{#1}

\newcommand{\const}[1]{#1}
\newcommand{\pred}[1]{\ensuremath{\mathsf{#1}}}
\newcommand{\true}{\ensuremath{\mathsf{True}}}
\newcommand{\false}{\ensuremath{\mathsf{False}}}
\newcommand{\rules}{\ensuremath{\mathcal{R}}}
\newcommand{\rulesplus}{\ensuremath{\mathcal{R^D}}}
\newcommand{\predsyms}[1]{\ensuremath{\mathcal{F}{(#1)}}}

\newcommand{\wmc}{\ensuremath{\mathrm{WMC}}}

\newcommand{\fig}[1]{Fig~\ref{#1}}

\usepackage{amsmath}
\usepackage{amssymb}
\usepackage{amsthm}

\newtheorem{proposition}{Proposition}
\newtheorem{defn}{Definition}
\newenvironment{definition}{\begin{defn}\em}{\end{defn}}
\newtheorem{exampl}{Example}
\newenvironment{example}{\begin{exampl}\em}{\end{exampl}}
\newtheorem{analog}{Analogy}

\newtheorem{motiv}{Motivation}

\newtheorem{conj}{Conjecture}
\newenvironment{conjecture}{\begin{conj}\em}{\end{conj}}

\newcommand{\citet}[1]{\citeauthor{#1} (\citeyear{#1})}

\nocopyright

\begin{document}

\title{Domain Recursion for Lifted Inference with Existential Quantifiers}
\author{ Seyed Mehran Kazemi$^1$, Angelika Kimmig$^2$, Guy Van den Broeck$^3$ \and David Poole$^1$ \\[0.2cm]
  $^\text{1}$ University of British Columbia, \texttt{\{smkazemi,poole\}@cs.ubc.ca}\\
  $^\text{2}$ KU Leuven, \texttt{angelika.kimmig@cs.kuleuven.be} \\
  $^\text{3}$ University of California, Los Angeles,  \texttt{guyvdb@cs.ucla.edu}
}
\maketitle
\begin{abstract}
In recent work, we proved that the domain recursion inference rule makes domain-lifted inference possible on several relational probability models (RPMs) for which the best known time complexity used to be exponential. 
We also identified two classes of RPMs for which inference becomes domain lifted when using domain recursion. These two classes subsume the largest lifted classes that were previously known.
In this paper, we show that domain recursion can also be applied to models with existential quantifiers.
Currently, all lifted inference algorithms assume that existential quantifiers have been removed in pre-processing by Skolemization. We show that besides introducing potentially inconvenient negative weights, Skolemization may increase the time complexity of inference. We give two example models where domain recursion can replace Skolemization, avoids the need for dealing with negative numbers, and reduces the time complexity of inference. These two examples may be interesting from three theoretical aspects: 1- they provide a better and deeper understanding of domain recursion and, in general, (lifted) inference, 2- they may serve as evidence that there are larger classes of models for which domain recursion can satisfyingly replace Skolemization, and 3- they may serve as evidence that better Skolemization techniques exist.

\end{abstract}

\section{Introduction}

\addtocounter{footnote}{3}

Statistical relational artificial intelligence (StarAI) \cite{StarAI-Book} aims at unifying logic and probability for reasoning and learning in noisy domains, described in terms of objects and the relationships among them. Relational probability models (RPMs) \cite{getoor2007introduction}, or template-based models \cite{Koller:2009}, are the core of StarAI. During the past decade and more, many RPMs have been developed \cite{Richardson:2006aa,DeRaedt:2007,Neville:2007,Poole:2008,PSL,natarajan2012gradient,Kazemi:2014}. A recent survey of these models can be found in \citet{kimmig2015lifted}. These models typically lift existing machine learning tools to succinctly represent probabilistic dependencies among properties and relationships of objects and provide compact representations of learned models.

One challenge with RPMs is the fact that they represent highly intractable, densely connected graphical models, typically with millions of random variables. In these models, objects about which there exists the same information are considered exchangeable. Exploiting such exchangeability is key for tractable inference and learning in RPMs~\cite{NiepertAAAI14}. Reasoning by exploiting the exchangeability among objects is known as \emph{lifted inference}. Lifted inference was first explicitly proposed by \citet{Poole:2003}. Since then, several researchers have worked on proposing rules providing exponential speedups for specific RPMs~\cite{De:2005,Milch:2008,Poole:2011,Choi:2011,jha2010lifted,PTP,Van:2011,van2014Skolemization}, speeding up lifted inference using the existing rules~\cite{Kazemi:2014a,LRC2CPP,kazemi2016compiling}, lifted learning of RPMs~\cite{ahmadi2012lifted,van2015lifted}, approximate lifted inference and learning~\cite{singla2008lifted,kersting2009counting,niepert2012markov,bui2013automorphism,VenugopalNIPS14,kopp2015lifted,jernite2015fast,anand2017non}, and characterizing the classes of RPMs that are amenable to efficient lifted inference \cite{gvdb2011completeness,taghipour2013completeness,beame2015symmetric}.

Characterizing the classes of RPMs that are amenable to efficient lifted inference began with the complexity notion of \emph{domain-lifted} inference~\cite{gvdb2011completeness} (a concept similar to data complexity in databases). Inference is domain-lifted when it runs in time polynomial in the number of objects in the domain.
In \citet{kazemi2016new}, we revived a forgotten lifted inference rule called \emph{domain recursion} and proved that for several RPMs, domain recursion makes domain-lifted inference possible. Examples include the \emph{symmetric transitivity} theory, the \emph{S4} clause of \citet{beame2015symmetric}, and a first-order formulation of the \emph{birthday paradox} \cite{birthday-paradox}. For all these theories, the previous best known time complexities using general purpose rules was exponential in the number of objects. We further identified two new classes $\StwoFOtwo$ and $\StwoRU$ of domain-liftable theories, which respectively subsume $\FOtwo$ and \emph{recursively unary}, the largest classes of domain-liftable theories known to that date. We also showed that even when domain recursion does not offer domain-lifted inference for an RPM, it may still reduce the complexity of inference exponentially.

This paper extends our results from \citet{kazemi2016new} by studying the applicability of the domain recursion rule for RPMs having existential quantifiers. Currently, if an RPM contains existential quantifiers, all existing software (including PTP\footnote{Available in Alchemy-2 \cite{Alchemy}}, WFOMC\footnote{\url{https://dtai.cs.kuleuven.be/software/wfomc}}, and L2C\footnote{\url{https://github.com/Mehran-k/L2C}}) either ground the existentially quantified variable, or remove existential quantifiers in a preprocessing step called Skolemization \cite{van2014Skolemization}. Grounding may increase the complexity of inference exponentially, so Skolemization is preferred over grounding except when there are only few objects. One known issue with Skolemization is that it introduces negative weights that are difficult to deal with in the implementation, as lifted inference computations are typically done in log space.

We identify two theories with existential quantifiers to which domain recursion can be applied without Skolemization. The first theory corresponds to the classic deck of cards problem from \citet{VdBKRR15}. The second theory is a reformulation of the deck of cards problem. For the first theory, domain recursion offers the same time complexity as Skolemization (quadratic in the number of objects), thus the only benefit of domain recursion is in avoiding to introduce negative weights. For the second one, besides avoiding negative weights, domain recursion offers a linear time complexity in the number of objects, whereas the complexity of inference in the Skolemized model is cubic. These theories offers three results. First, they show that Skolemization may increase the time complexity of lifted inference (which has not been previously observed). Second, they show that domain recursion (when applicable) may be a suitable replacement for Skolemization. Third, they show that reformulation of the theory for a problem can potentially reduce its time complexity. These results will be established using the weighted model counting (\wmc) formulation of RPMs~\cite{Van:2011}.

\section{Background \& Notation}
In this section, we will introduce our notation and provide necessary background.

\subsection{Finite-domain, function-free first-order logic}
A \textbf{population} is a finite set of objects (or individuals). 
A \textbf{logical variable (logvar)} is typed with a population. 
We represent logvars with lower-case letters. 
The population associated with a logvar $\lv{x}$ is $\pop{x}$. The cardinality of $\pop{x}$ is $|\pop{x}|$. 
For every object, we assume there exists a unique \emph{constant} denoting that object. Thus we are making the \emph{unique names assumption}, where different constants (names) refer to different objects. A lower-case letter in bold represents a tuple of logvars and an upper-case letter in bold represents a tuple of constants. We use $\lv{x}\in\pop{x}$ as a shorthand for instantiating $\lv{x}$ with one of the elements of $\pop{x}$.

An \textbf{atom} is of the form $\pred{F}(t_1, \dots, t_k)$ where $\pred{F}$ is a predicate symbol and each $t_i$ is a logvar or a constant. A \textbf{unary} atom contains exactly one logvar and a \textbf{binary} atom contains exactly two logvars. 
A \textbf{grounding} of an atom is obtained by replacing each of its logvars $\lv{x}$ by one of the objects in $\pop{x}$.
A \textbf{literal} is an atom or its negation.
A \textbf{formula} $\varphi$ is a literal, a disjunction $\varphi_1\vee\varphi_2$ of formulas, a conjunction $\varphi_1\wedge\varphi_2$ of formulas, or a quantified formula $\forall\lv{x}\in\pop{x}:\varphi(\lv{x})$ or $\exists\lv{x}\in\pop{x}:\varphi(\lv{x})$ where $\lv{x}$ appears in $\varphi(\lv{x})$.
A formula is in \textbf{prenex normal form} if it is written as a string of quantifiers followed by a quantifier-free part.
A \textbf{sentence} is a formula with all logvars quantified. 
A \textbf{clause} is a disjunction of literals.
A \textbf{theory} is a set of sentences.
A theory is \textbf{clausal} if all its sentences are clauses. 
In this paper, we assume all theories are clausal and the clauses are in prenex normal form. When a clause mentions two logvars $\lv{x_1}$ and $\lv{x_2}$ typed with the same population $\pop{x}$, or a logvar $\lv{x}$ typed with population $\pop{x}$ and a constant~$\const{C}\in\pop{x}$, we assume they refer to different objects\footnote{Equivalently, we can disjoin $\lv{x_1} = \lv{x_2}$ or $\lv{x} = \const{C}$ to the clause. Note that any theory without this restriction can be transformed into a theory conforming to this restriction.}.

An \textbf{interpretation} is an assignment of values to all ground atoms in a theory. 
An interpretation~$I$ is a \textbf{model} of a theory~$T$, $I\models T$, if given its value assignments, all sentences in $T$ evaluate to \true.

\subsection{Weighted Model Counting}
Let $\predsyms{T}$ be the set of predicate symbols in theory $T$, and $\Phi: \predsyms{T}\rightarrow \mathbb{R}$ and $\overline{\Phi}: \predsyms{T}\rightarrow \mathbb{R}$ be two functions 
that map each predicate~$\pred{S}$ to weights.
These functions associate a weight with assigning \true\ or \false\ to groundings~$\pred{S}(\const{C}_1,\ldots,\const{C}_k)$.
Let $I$ be an interpretation for $T$, $\kappa^{True}$ be the set of groundings assigned \true\ in $I$, and $\kappa^{False}$ the ones assigned \false. The weight of $I$ is given by:
\begin{equation}
w(I)=\prod_{ \pred{S}(\const{C}_1,\ldots,\const{C}_k)\in \kappa^{True}} \Phi(\pred{S}) \cdot \prod_{\pred{S}(\const{C}_1,\ldots,\const{C}_k) \in \kappa^{False}} \overline{\Phi}(\pred{S})
\end{equation} 
Given a theory $T$ and two functions $\Phi$ and $\overline{\Phi}$, the \textbf{weighted model count (WMC)} of the theory given $\Phi$ and $\overline{\Phi}$ is:
\begin{equation}
\wmc(T | \Phi, \overline{\Phi}) = \sum_{I \models T} w(I).
\end{equation}

\begin{example}
Consider the theory: 
\begin{align*}
\forall \lv{x} \in \pop{x}: \neg \pred{Young}(\lv{x}) \vee \pred{Adventurous}(\lv{x})
\end{align*}
having only one clause and assume $\pop{x} = \{\const{A}, \const{B}\}$. 
The truth assignment $\pred{Young}(\const{A})=\false, \pred{Young}(\const{B})=\true, \pred{Adventurous}(\const{A})=\true, \pred{Adventurous}(\const{B})=\true$ is a model. 
Assuming $\Phi(\pred{Young})=0.3$, $\Phi(\pred{Adventurous})=0.7$, $\overline{\Phi}(\pred{Young})=0.4$ and $\overline{\Phi}(\pred{Adventurous})=1.1$, the weight of this model is $0.4 * 0.3 * 0.7 * 0.7$. 
This theory has eight other models. The \wmc\ can be calculated by summing the weights of all nine models.
\end{example}

\subsection{Skolemization} 
Consider a theory $T$ having a clause $C=\forall \lvs{x}, \exists \lv{y}:  \varphi$ where $\varphi$ is a formula containing $\lvs{x}$, $\lv{y}$, and (possibly) other quantified logvars. Let $C'=\forall \lvs{x}, \forall \lv{y}:  \neg \varphi \vee \pred{A}(\lvs{x})$, where \pred{A} is a predicate not appearing in $T$. Let $T'$ be theory $T$ where $C$ is replaced with $C'$. Then \citet{van2014Skolemization} prove that for any weight functions $\Phi$ and $\overline{\Phi}$, $\wmc(T \mid \Phi, \overline{\Phi})=\wmc(T' \mid \Phi', \overline{\Phi'})$, where $\Phi'$ is $\Phi$ extended with a weight $1$ for $\pred{A}$, and $\overline{\Phi'}$ is $\overline{\Phi}$ extended with a weight $-1$ for $\neg \pred{A}$. A Skolemized theory (a theory with no existential quantifier) can be obtained by repeatedly applying this process until no more existential quantifiers exist.

\subsection{Calculating the WMC of a theory} 
Given a Skolemized theory $T$ as input, the \wmc\ of the theory can be calculated by applying a set \rules\ of rules. These rules include (lifted) decomposition, (lifted) case analysis, unit propagation, shattering, caching, grounding, and Skolemization. For the details of these rules, we direct readers to \citet{kazemi2016new}. Here we only explain symbolic application of lifted case analysis.

Let $T$ be a theory and $\pred{S}$ be a predicate in $T$ having only one logvar $\lv{x}$. Lifted case analysis on $\pred{S}$ generates $|\pop{x}|+1$ sub-theories where the $i$-th sub-theory corresponds to $\pred{S}$ being \true\ for $i$ out of $|\pop{x}|$ objects, and \false\ for the rest. Symbolic application of lifted case analysis on $\pred{S}$ generates only one sub-theory corresponding to $\pred{S}$ being \true\ for $i$ out of $|\pop{x}|$ objects, and \false\ for the rest, for a generic $i$. Symbolic application of lifted case analysis has been used for compilation purposes \cite{Van:2011,LRC2CPP}.

\subsection{Domain-Liftability}
A theory is \textbf{domain-liftable} \cite{gvdb2011completeness} if calculating its \wmc\ is $O(Poly(|\pop{x_1}|, |\pop{x_2}|, \dots, |\pop{x_k}|))$, where $\lv{x}_1, \lv{x}_2, \dots, \lv{x}_k$ represent the logvars in the theory. A class $C$ of theories is domain-liftable if every $T \in C$ is domain-liftable. 

\paragraph{$\mathbf{FO^2}$:} A theory is in $\FOtwo$ if all its clauses have up to two logvars. $\FOtwo$ is a domain-liftable class of RPMs~\cite{gvdb2011completeness,van2014Skolemization}.

\paragraph{Recursively unary (RU):} A theory $T$ is \emph{recursively unary (RU)} if for every theory $T'$ that results from exhaustively applying to $T$ all rules in \rules\ except for lifted case analysis, either one of the following holds:
\begin{itemize}
\item[-] $T'$ has no sentences, or
\item[-] there exists some atom $\pred{S}$ in $T'$ containing only one logvar \lv{x}, and the result of applying (symbolic) lifted case analysis on $\pred{S}$ is \emph{RU}.
\end{itemize}
\emph{RU} is a domain-liftable class of RPMs~\cite{Poole:2011}.

Note that membership checking for both $\FOtwo$ and $RU$ is independent of the population sizes.

\section{Domain recursion}
\textbf{Domain recursion} can be viewed as a grounding scheme which grounds only one object from a population at a time, and then applies the standard rules in \rules\ to the modified theory, until the grounded object is entirely removed from the theory. 
More formally, let \lv{x} be a logvar in a theory $T$, $\ind{N}$ be an object in $\pop{x}$, and $\pop{x'}=\pop{x}-\{\ind{N}\}$. Domain recursion rewrites the clauses in $T$ in terms of $\pop{x'}$ and $\ind{N}$ (i.e. grounding only one object from the population), removing $\pop{x}$ from the theory entirely.
Then, standard rules in \rules\ are applied to the modified theory until the grounded object ($\ind{N}$) is entirely removed from the theory. Note that when $\ind{N}$ is removed from the theory, the resulting theory may have different sentences compared to the original theory.

Domain recursion is \textbf{bounded} if after applying the standard rules in \rules\ to the modified theory until $\ind{N}$ is entirely removed from the theory, the problem is reduced to a \wmc\ problem on a theory identical to the original one, except that $\pop{x}$ is replaced by the smaller domain $\pop{x'}$. When domain recursion is bounded for a theory, we can compute its \wmc\ using dynamic programming. We refer to \rules\ extended with \emph{bounded domain recursion (BDR)} rule as \rulesplus.

\begin{example} \label{fxy_then_fyx}
Suppose we have a theory whose only clause is
\[\forall \lv{x}, \lv{y} \in \pop{p}: \neg \pred{MarriedTo}(\lv{x}, \lv{y}) \vee \pred{MarriedTo}(\lv{y}, \lv{x})\]
stating if $\lv{x}$ is married to $\lv{y}$, $\lv{y}$ is also married to $\lv{x}$. In order to calculate the \wmc\ of this theory using domain recursion, we let $\ind{N}$ be an object in $\pop{p}$, $\pop{p'}=\pop{p}-\{\ind{N}\}$, and re-write the theory in terms of $\pop{p'}$ and $\ind{N}$ as:
\begin{align*}
\forall \lv{x} \in \pop{p'}:&~ \neg \pred{MarriedTo}(\lv{x}, \ind{N}) \vee \pred{MarriedTo}(\ind{N}, \lv{x})\\
\forall \lv{y} \in \pop{p'}:&~ \neg \pred{MarriedTo}(\ind{N}, \lv{y}) \vee \pred{MarriedTo}(\lv{y}, \ind{N})\\
\forall \lv{x}, \lv{y} \in \pop{p'}:&~ \neg \pred{MarriedTo}(\lv{x}, \lv{y}) \vee \pred{MarriedTo}(\lv{y}, \lv{x})
\end{align*}
Performing lifted case analysis on $\pred{MarriedTo}(\lv{p'}, \ind{N})$ and $\pred{MarriedTo}(\ind{A}, \lv{p'})$, shattering and unit propagation gives
\[\forall \lv{x}, \lv{y} \in \pop{p'}: \neg \pred{MarriedTo}(\lv{x}, \lv{y}) \vee \pred{MarriedTo}(\lv{y}, \lv{x})\]
This theory is equivalent to the original theory, except that $\pop{p}$ is replaced with the smaller $\pop{p'}$. Therefore, domain recursion is bounded on this theory. By keeping a cache of the values of each sub-theory, one can verify that BDR finds the \wmc\ of the above theory in polynomial time.
\end{example}

\section{Summary of \citet{kazemi2016new}}
While domain recursion was central to the proof that $\FOtwo$ is domain-liftable, later work showed that simpler rules suffice to capture $\FOtwo$~\cite{taghipour2013completeness}, and the domain recursion rule was forgotten. In \citet{kazemi2016new}, we revived the domain recursion rule by showing that it makes more theories domain-liftable. Using \rulesplus, we identified new classes $\StwoFOtwo$ and $\StwoRU$ of domain-liftable theories, which respectively subsume $\FOtwo$ and \emph{RU}. 

\begin{definition} \label{s2fo2-def}
Let $\alpha(\pred{S})$ be a clausal theory that uses a single binary predicate $\pred{S}$, such that each clause has exactly two different literals of $\pred{S}$. 
Let $\alpha = \alpha(\pred{S}_1) \wedge \alpha(\pred{S}_2) \wedge \dots \wedge \alpha(\pred{S}_n)$ where the $\pred{S}_i$ are different binary predicates. 
Let $\beta$ be a theory where all clauses contain at most one $\pred{S}_i$ literal, and the clauses that contain an $\pred{S}_i$ literal contain no other literals with more than one logvar. 
Then, $\StwoFOtwo$ and $\StwoRU$ are the classes of theories of the form $\alpha \wedge \beta$ where $\beta \in \FOtwo$ and $\beta \in RU$ respectively.
\end{definition}

\citet{beame2015symmetric} identified the following clause:
\begin{align*}
\forall \lv{x}_1, \lv{x}_2 \in &~\pop{x}, \lv{y}_1, \lv{y}_2 \in \pop{y}: \\
&~\pred{S}(\lv{x}_1, \lv{y}_1) \vee \neg \pred{S}(\lv{x}_2, \lv{y}_1) \vee \pred{S}(\lv{x}_2, \lv{y}_2) \vee \neg \pred{S}(\lv{x}_1, \lv{y}_2)
\end{align*}
 and proved that, even though the rules in \rules\ cannot calculate the \wmc\ of that clause, there is a polynomial-time algorithm for finding its \wmc. They concluded that this set of rules \rules\ for finding the \wmc\ of theories does not suffice, asking for new rules to compute their theory. We proved that adding bounded domain recursion to the set achieves this goal.

Domain-liftable calculation of \wmc\ for the transitivity formula is a long-standing open problem. Symmetric transitivity containing the following clauses:
\begin{align*}
\forall \lv{x}, \lv{y}, \lv{z} \in \pop{p}:&~ \neg \pred{F}(\lv{x}, \lv{y}) \vee \neg \pred{F}(\lv{y}, \lv{z}) \vee \pred{F}(\lv{x}, \lv{z})\\
\forall \lv{x}, \lv{y} \in \pop{p}:&~ \neg \pred{F}(\lv{x}, \lv{y}) \vee \pred{F}(\lv{y}, \lv{x})
\end{align*} 
is easier as its model count corresponds to the Bell number, but solving it using general-purpose rules has been an open problem. We proved that the rules in \rulesplus\ suffice for solving symmetric transitivity using general-purpose rules.

In all the above theories, domain recursion is bounded: after revealing one object and re-writing the theory in terms of that object and the reduced domain, we are able to apply the rules \rules\ to the modified theory and get back the original theory over a reduced domain. For theories where domain recursion is not bounded, we showed that is may still offer exponential speedup compared to grounding all objects of a population simultaneously.

\subsection{BDR+RU Class}
\begin{figure}
\begin{center}
\includegraphics[width=\columnwidth]{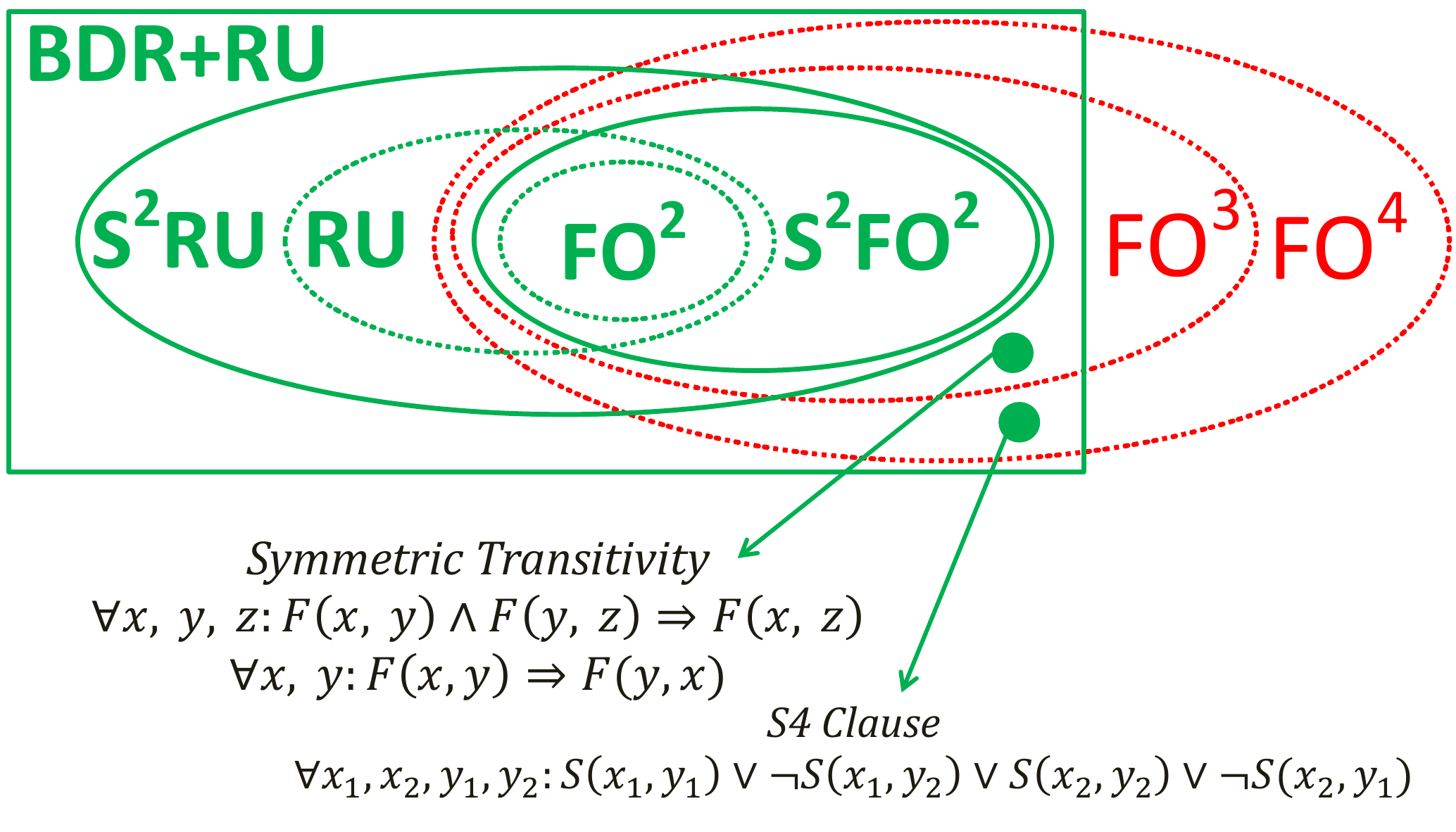}
\end{center}
\caption{A summary of the domain-liftability results. $\FOtwo$ and \emph{RU} have been previously proved to be domain-liftable. For $\mathit{FO}^3$, it has been proved that at least one of its theories is not domain-liftable under standard complexity assumptions. We identified the new domain-liftable classes $\StwoFOtwo$ and $\StwoRU$ and proved that $\FOtwo \subset \StwoFOtwo$, $RU \subset \StwoRU$, $\FOtwo \subset RU$, and $\StwoFOtwo \subset \StwoRU$. We also proved that \emph{symmetric transitivity} and the \emph{S4 clause} are domain-liftable. The \textbf{\textit{BDR+RU}} class introduced in this paper contains all theories that are currently known to be domain-liftable.}
\label{liftability-results}
\end{figure}

Let \textbf{\textit{BDR+RU}} be the class of theories defined identically as $RU$, except that \rules\ is replaced with \rulesplus\ in the definition. Note that similar to $RU$, membership checking for \textbf{\textit{BDR+RU}} is also independent of the population sizes.
Our new domain-liftability results, the previously known results, and the newly introduced can be summarized as in \fig{liftability-results}. Currently, \textbf{\textit{BDR+RU}} contains all theories for which we know an algorithm for polynomial-time computation of their \wmc\ exists. 
This leads us to conjecture the following.
\begin{conjecture}
\textbf{\textit{BDR+RU}} is the largest possible class of domain-liftable RPMs.
\end{conjecture}
This conjecture can be also stated as follows.
\begin{conjecture}
\rulesplus\ is a complete set of rules for domain-lifted inference.
\end{conjecture}

\section{Inference with Existential Quantifiers}
In this section, we introduce two theories having existential quantifiers for which the \wmc\ can be calculated using domain recursion without Skolemizing. 

\paragraph{Deck of cards problem:} \citet{VdBKRR15} introduced a problem known as the deck of cards problem to show how lifted inference can be exponentially faster than classic inference. Consider a deck of cards where all 52 cards are faced down on the ground, and assume we need to answer queries such as the probability of the first card being the \emph{queen of hearts (QofH)}, the probability of the second card being a \emph{hearts} given that the first card is the \emph{QofH}, etc. 
Let $\pop{c}$ be a population whose objects are the cards in a deck of cards, and $\pop{p}$ represent the possible positions for these cards. Suppose $|\pop{c}|=|\pop{p}|=n(=52)$. The following theory corresponds to \citet{VdBKRR15}'s formulation of the deck of cards problem. It states that for each card, there exists a position. For each position there exists a card. And no two cards can be in the same position.
\begin{align*}
\forall \lv{c} \in \pop{c}, \exists \lv{p} \in \pop{p}:&~ \pred{S}(\lv{c}, \lv{p}) \\
\forall \lv{p} \in \pop{p}, \exists \lv{c} \in \pop{c}:&~ \pred{S}(\lv{c}, \lv{p})\\
\forall \lv{p} \in \pop{p}, \lv{c}_1, \lv{c}_2 \in \pop{c}:&~ \neg \pred{S}(\lv{c}_1, \lv{p}) \vee \neg \pred{S}(\lv{c}_2, \lv{p})
\end{align*}
\begin{proposition} \label{exists1-prop}
The \wmc\ of the theory for \citet{VdBKRR15}'s deck of cards problem can be calculated in $O(n^2)$ using the rules in \rulesplus\ without Skolemization.
\end{proposition}

For the theory in Proposition~\ref{exists1-prop}, we could alternatively apply Skolemization on the theory and then solve the Skolemized theory. The Skolemized theory will be as follows:
\begin{align*}
\forall \lv{c} \in \pop{c}, \lv{p} \in \pop{p}:&~ \neg \pred{S}(\lv{c}, \lv{p}) \vee \pred{A}(\lv{c}) \\
\forall \lv{c} \in \pop{c}, \lv{p} \in \pop{p}:&~ \neg \pred{S}(\lv{c}, \lv{p}) \vee \pred{B}(\lv{p}) \\
\forall \lv{p} \in \pop{p}, \lv{c}_1, \lv{c}_2 \in \pop{c}:&~ \neg \pred{S}(\lv{c}_1, \lv{p}) \vee \neg \pred{S}(\lv{c}_2, \lv{p})
\end{align*}
where $\pred{A}(\lv{c})$ and $\pred{B}(\lv{p})$ are the literals added to the theory as a result of Skolemization. One can verify that the time complexity of this alternative is $O(n^2)$ as well\footnote{In order to find the \wmc\ of the theory, lifted case analysis can be applied on $\pred{A}(\lv{c})$ and $\pred{B}(\lv{p})$, then unit propagation can be applied on unit clauses resulting from the first two clauses, then lifted decomposition can be applied on $\lv{p}$ of the third clause, and the \wmc\ of the resulting theory is obvious. The time complexity is $O(n^2)$ due to the two lifted case-analyses.}. However, by applying domain recursion directly to the non-Skolemized theory, we avoid dealing with negative weights in the implementation. Note that dealing with negative weights can be difficult as the computations for \wmc\ are typically done in log space.

\paragraph{A reformulation of the deck of cards problem:} The following theory is a reformulation of the deck of cards problem. It contains all sentences from \citet{VdBKRR15}, plus a sentence stating a position cannot belong to more than one card. Semantically, this theory is identical to \citet{VdBKRR15}'s theory.
\begin{align*}
\forall \lv{c} \in \pop{c}, \exists \lv{p} \in \pop{p}:&~ \pred{S}(\lv{c}, \lv{p})\\
\forall \lv{p} \in \pop{p}, \exists \lv{c} \in \pop{c}:&~ \pred{S}(\lv{c}, \lv{p})\\
\forall \lv{p} \in \pop{p}, \lv{c}_1, \lv{c}_2 \in \pop{c}:&~ \neg \pred{S}(\lv{c}_1, \lv{p}) \vee \neg \pred{S}(\lv{c}_2, \lv{p}) \\
\forall \lv{p}_1, \lv{p}_2 \in \pop{p}, \lv{c} \in \pop{c}:&~ \neg \pred{S}(\lv{c}, \lv{p}_1) \vee \neg \pred{S}(\lv{c}, \lv{p}_2)
\end{align*}

\begin{proposition} \label{exists2-prop}
The \wmc\ of the reformulated theory for the deck of cards problem can be calculated in $O(n)$ using the rules in \rulesplus\ without Skolemization.
\end{proposition}

For the theory in Proposition~\ref{exists2-prop}, we could alternatively apply Skolemization on the theory and then solve the Skolemized theory. The Skolemized theory will be as follows:
\begin{align*}
\forall \lv{c} \in \pop{c}, \lv{p} \in \pop{p}:&~ \neg \pred{S}(\lv{c}, \lv{p}) \vee \pred{A}(\lv{c}) \\
\forall \lv{c} \in \pop{c}, \lv{p} \in \pop{p}:&~ \neg \pred{S}(\lv{c}, \lv{p}) \vee \pred{B}(\lv{p}) \\
\forall \lv{p} \in \pop{p}, \lv{c}_1, \lv{c}_2 \in \pop{c}:&~ \neg \pred{S}(\lv{c}_1, \lv{p}) \vee \neg \pred{S}(\lv{c}_2, \lv{p}) \\
\forall \lv{p}_1, \lv{p}_2 \in \pop{p}, \lv{c} \in \pop{c}:&~ \neg \pred{S}(\lv{c}, \lv{p}_1) \vee \neg \pred{S}(\lv{c}, \lv{p}_2)
\end{align*}
where $\pred{A}(\lv{c})$ and $\pred{B}(\lv{p})$ are the literals added to the theory as a result of Skolemization. One can verify that the time complexity of this alternative is $O(n^3)$ instead of linear\footnote{In order to find the \wmc\ of the theory, lifted case analysis can be applied on $\pred{A}(\lv{c})$ and $\pred{B}(\lv{p})$, then unit propagation can be applied on unit clauses resulting from the first two clauses, then domain recursion can solve the two remaining clauses. The time complexity is $O(n^3)$ because there are two lifted case-analyses and then an $O(n)$ domain recursion step.}. This shows that beside introducing negative weights that are difficult to deal with in the implementation, Skolemization may also increase the time complexity of \wmc\ for some theories. Therefore, domain recursion (when applicable) may be a better option compared to Skolemization for theories with existential quantifiers.

\section{Conclusion and Future Work}
In this paper, we summarized our new domain-liftability results using domain recursion \cite{kazemi2016new}. Then we extended our results by showing that domain recursion can be also applied to the theories with existential quantifiers. On two example theories, we showed that the previous approach for handling existential quantifiers (i.e. Skolemization) introduces negative weights to the theory that are difficult to deal with in the implementation, and may increase the time complexity. We showed that domain recursion can be a suitable replacement for both these theories. 

While we have so far identified several theories (or classes of theories) with and without existential quantifier for which domain recursion is bounded, it is still unclear what property of a theory makes it amenable to bounded domain recursion (BDR). In other words, the only way we can currently know whether domain recursion is bounded for a theory or not is by applying domain recursion to the theory. This opens two avenues for future work: 1- identifying the properties making theories amenable to BDR , 2- characterizing more or larger classes of theories for which domain recursion is bounded.

\appendix
\setcounter{secnumdepth}{0}
\section{Appendix}
\textit{Proof of Proposition~\ref{exists1-prop}.}
Let $\ind{N}$ be an object in $\pop{p}$ and $\pop{p'} = \pop{p}-\{\ind{N}\}$. We can re-write the theory in terms of $\ind{N}$ and $\pop{p'}$ as follows:
\[\forall \lv{c} \in \pop{c}: \pred{S}(\lv{c}, \ind{N}) \vee (\exists \lv{p'} \in \pop{p'}: \pred{S}(\lv{c}, \lv{p'}))\]
\[\exists \lv{c} \in \pop{c}: \pred{S}(\lv{c}, \ind{N})\]
\[\forall \lv{p'} \in \pop{p'}, \exists \lv{c} \in \pop{c}: \pred{S}(\lv{c}, \lv{p'})\]
\[\forall \lv{c}_1, \lv{c}_2 \in \pop{c}: \neg \pred{S}(\lv{c}_1, \ind{N}) \vee \neg \pred{S}(\lv{c}_2, \ind{N})\]
\[\forall \lv{p'} \in \pop{p'}, \lv{c}_1, \lv{c}_2 \in \pop{c}: \neg \pred{S}(\lv{c}_1, \lv{p'}) \vee \neg \pred{S}(\lv{c}_2, \lv{p'})\]
Since $\exists \lv{c} \in \pop{c}: \pred{S}(\lv{c}, \ind{N})$ (the second clause above), let $\ind{M}$ be an object in $\pop{c}$ such that $\pred{S}(\ind{M}, \ind{N})=\true$ and let $\pop{c'} = \pop{c} - \{\ind{M}\}$. By re-writing the theory in terms of $\pop{c'}$ and $\ind{M}$, the second clause can be removed and we will get the following theory\footnote{Note that there are $|\pop{c}|$ ways for selecting $\ind{M}$, so the \wmc\ of the resulting theory must be multiplied by $|\pop{c}|$. Also note that if lifted case analysis is subsequently performed on $\pred{S}(\lv{c}, \ind{N})$, the \wmc\ of the $i$-th branch must be divided by $i+1$. \label{footnote1}}:
\[\pred{S}(\ind{M}, \ind{N}) \vee (\exists \lv{p'} \in \pop{p'}: \pred{S}(\ind{M}, \lv{p'}))\]
\[\forall \lv{c'} \in \pop{c'}: \pred{S}(\lv{c}, \ind{N}) \vee (\exists \lv{p'} \in \pop{p'}: \pred{S}(\lv{c}, \lv{p'}))\]
\[\forall \lv{p'} \in \pop{p'}: \pred{S}(\ind{M}, \lv{p'}) \vee (\exists \lv{c'} \in \pop{c'}: \pred{S}(\lv{c}, \lv{p'}))\]
\[\forall \lv{c'}_1 \in \pop{c'}: \neg \pred{S}(\lv{c'}_1, \ind{N}) \vee \neg \pred{S}(\ind{M}, \ind{N})\]
\[\forall \lv{c'}_1, \lv{c'}_2 \in \pop{c'}: \neg \pred{S}(\lv{c'}_1, \ind{N}) \vee \neg \pred{S}(\lv{c'}_2, \ind{N})\]
\[\forall \lv{p'} \in \pop{p'}, \lv{c'}_1 \in \pop{c'}: \neg \pred{S}(\lv{c'}_1, \lv{p'}) \vee \neg \pred{S}(\ind{M}, \lv{p'})\]
\[\forall \lv{p'} \in \pop{p'}, \lv{c'}_1, \lv{c'}_2 \in \pop{c'}: \neg \pred{S}(\lv{c'}_1, \lv{p'}) \vee \neg \pred{S}(\lv{c'}_2, \lv{p'})\]
The first clause can be removed as we know $\pred{S}(\ind{M}, \ind{N})$ is \true. The fourth clause can be simplified to $\forall \lv{c'}_1 \in \pop{c'}: \neg \pred{S}(\lv{c'}_1, \ind{N})$. Since this clause now contains only one literal, we can apply unit propagation on it. Unit propagation on $\pred{S}(\lv{c'}_1, \ind{N})$ will give the following theory:
\[\forall \lv{c'} \in \pop{c'}, \exists \lv{p'} \in \pop{p'}: \pred{S}(\lv{c}, \lv{p'})\]
\[\forall \lv{p'} \in \pop{p'}: \pred{S}(\ind{M}, \lv{p'}) \vee (\exists \lv{c'} \in \pop{c'}: \pred{S}(\lv{c}, \lv{p'}))\]
\[\forall \lv{p'} \in \pop{p'}, \lv{c'}_1 \in \pop{c'}: \neg \pred{S}(\lv{c'}_1, \lv{p'}) \vee \neg \pred{S}(\ind{M}, \lv{p'})\]
\[\forall \lv{p'} \in \pop{p'}, \lv{c'}_1, \lv{c'}_2 \in \pop{c'}: \neg \pred{S}(\lv{c'}_1, \lv{p'}) \vee \neg \pred{S}(\lv{c'}_2, \lv{p'})\]
Lifted case analysis on $\pred{S}(\ind{M}, \lv{p'})$ assuming $\pop{p'_T}$ and $\pop{p'_F}$ represent mutual exclusive and covering subsets of $\pop{p'}$ such that $\forall \lv{p'}_T \in \pop{p'_T}: \pred{S}(\ind{M}, \lv{p'}_T)$ and $\forall \lv{p'}_F \in \pop{p'_F}: \neg \pred{S}(\ind{M}, \lv{p'}_F)$ will give the following theory:
\[\forall \lv{c'} \in \pop{c'}: (\exists \lv{p'}_T \in \pop{p'_T}: \pred{S}(\lv{c}, \lv{p'}_T) \vee \exists \lv{p'}_F \in \pop{p'_F}: \pred{S}(\lv{c}, \lv{p'}_F))\]
\[\forall \lv{p'}_F \in \pop{p'_F} \exists \lv{c'} \in \pop{c'}: \pred{S}(\lv{c}, \lv{p'}_F)\]
\[\forall \lv{p'}_T \in \pop{p'_T}, \lv{c'}_1 \in \pop{c'}: \neg \pred{S}(\lv{c'}_1, \lv{p'}_T)\]
\[\forall \lv{p'}_T \in \pop{p'_T}, \lv{c'}_1, \lv{c'}_2 \in \pop{c'}: \neg \pred{S}(\lv{c'}_1, \lv{p'}_T) \vee \neg \pred{S}(\lv{c'}_2, \lv{p'}_T)\]
\[\forall \lv{p'}_F \in \pop{p'_F}, \lv{c'}_1, \lv{c'}_2 \in \pop{c'}: \neg \pred{S}(\lv{c'}_1, \lv{p'}_F) \vee \neg \pred{S}(\lv{c'}_2, \lv{p'}_F)\]
Unit propagation on the third clause gives the following theory:
\[\forall \lv{c'} \in \pop{c'} \exists \lv{p'}_F \in \pop{p'_F}: \pred{S}(\lv{c}, \lv{p'}_F)\]
\[\forall \lv{p'}_F \in \pop{p'_F} \exists \lv{c'} \in \pop{c'}: \pred{S}(\lv{c}, \lv{p'}_F)\]
\[\forall \lv{p'}_F \in \pop{p'_F}, \lv{c'}_1, \lv{c'}_2 \in \pop{c'}: \neg \pred{S}(\lv{c'}_1, \lv{p'}_F) \vee \neg \pred{S}(\lv{c'}_2, \lv{p'}_F)\]
This theory is identical to the original theory that we started with, except that it is over the smaller populations $\pop{c'}$ and $\pop{p'_F}$. By continuing this process and keeping the intermediate results in a cache, the \wmc\ of the theory can be found in $O(n^2)$: the number of times we do the above procedure is at most $n=|\pop{c}|$ and each time we do the procedure it takes $O(n=|\pop{p}|)$ time for the lifted case analysis. \hfill$\qedsymbol$ \\

\textit{Proof of Proposition~\ref{exists2-prop}.}
Let $\ind{N}$ be an object in $\pop{p}$ and $\pop{p'} = \pop{p}-\{\ind{N}\}$. We can re-write the theory in terms of $\ind{N}$ and $\pop{p'}$ as follows:
\[\forall \lv{c} \in \pop{c}: \pred{S}(\lv{c}, \ind{N}) \vee (\exists \lv{p'} \in \pop{p'}: \pred{S}(\lv{c}, \lv{p'}))\]
\[\exists \lv{c} \in \pop{c}: \pred{S}(\lv{c}, \ind{N})\]
\[\forall \lv{p'} \in \pop{p'}, \exists \lv{c} \in \pop{c}: \pred{S}(\lv{c}, \lv{p'})\]
\[\forall \lv{c}_1, \lv{c}_2 \in \pop{c}: \neg \pred{S}(\lv{c}_1, \ind{N}) \vee \neg \pred{S}(\lv{c}_2, \ind{N})\]
\[\forall \lv{p'} \in \pop{p'}, \lv{c}_1, \lv{c}_2 \in \pop{c}: \neg \pred{S}(\lv{c}_1, \lv{p'}) \vee \neg \pred{S}(\lv{c}_2, \lv{p'})\]
\[\forall \lv{p'}_1 \in \pop{p'}, \lv{c} \in \pop{c}: \neg \pred{S}(\lv{c}, \lv{p'}_1) \vee \neg \pred{S}(\lv{c}, \ind{N})\]
\[\forall \lv{p'}_1, \lv{p'}_2 \in \pop{p'}, \lv{c} \in \pop{c}: \neg \pred{S}(\lv{c}, \lv{p'}_1) \vee \neg \pred{S}(\lv{c}, \lv{p'}_2)\]

Since $\exists \lv{c} \in \pop{c}: \pred{S}(\lv{c}, \ind{N})$ (the second clause above), let $\ind{M}$ be an object in $\pop{c}$ such that $\pred{S}(\ind{M}, \ind{N})$ and let $\pop{c'} = \pop{c} - \{\ind{M}\}$. By re-writing the theory in terms of $\pop{c'}$ and $\ind{M}$, the second clause can be removed and we will get the following theory\footnote{cf. footnote~\ref{footnote1}}:
\[\pred{S}(\ind{M}, \ind{N}) \vee (\exists \lv{p'} \in \pop{p'}: \pred{S}(\ind{M}, \lv{p'}))\]
\[\forall \lv{c'} \in \pop{c'}: \pred{S}(\lv{c}, \ind{N}) \vee (\exists \lv{p'} \in \pop{p'}: \pred{S}(\lv{c}, \lv{p'}))\]
\[\forall \lv{p'} \in \pop{p'}: \pred{S}(\ind{M}, \lv{p'}) \vee (\exists \lv{c'} \in \pop{c'}: \pred{S}(\lv{c}, \lv{p'}))\]
\[\forall \lv{c'}_1 \in \pop{c'}: \neg \pred{S}(\lv{c'}_1, \ind{N}) \vee \neg \pred{S}(\ind{M}, \ind{N})\]
\[\forall \lv{c'}_1, \lv{c'}_2 \in \pop{c'}: \neg \pred{S}(\lv{c'}_1, \ind{N}) \vee \neg \pred{S}(\lv{c'}_2, \ind{N})\]
\[\forall \lv{p'} \in \pop{p'}, \lv{c'}_1 \in \pop{c'}: \neg \pred{S}(\lv{c'}_1, \lv{p'}) \vee \neg \pred{S}(\ind{M}, \lv{p'})\]
\[\forall \lv{p'} \in \pop{p'}, \lv{c'}_1, \lv{c'}_2 \in \pop{c'}: \neg \pred{S}(\lv{c'}_1, \lv{p'}) \vee \neg \pred{S}(\lv{c'}_2, \lv{p'})\]
\[\forall \lv{p'}_1 \in \pop{p'}: \neg \pred{S}(\ind{M}, \lv{p'}_1) \vee \neg \pred{S}(\ind{M}, \ind{N})\]
\[\forall \lv{p'}_1 \in \pop{p'}, \lv{c'} \in \pop{c'}: \neg \pred{S}(\lv{c'}, \lv{p'}_1) \vee \neg \pred{S}(\lv{c'}, \ind{N})\]
\[\forall \lv{p'}_1, \lv{p'}_2 \in \pop{p'}: \neg \pred{S}(\ind{M}, \lv{p'}_1) \vee \neg \pred{S}(\ind{M}, \lv{p'}_2)\]
\[\forall \lv{p'}_1, \lv{p'}_2 \in \pop{p'}, \lv{c'} \in \pop{c'}: \neg \pred{S}(\lv{c'}, \lv{p'}_1) \vee \neg \pred{S}(\lv{c'}, \lv{p'}_2)\]

The first clause can be removed as we know $\pred{S}(\ind{M}, \ind{N})=\true$. The fourth and the eighth clauses can be simplified to $\forall \lv{c'}_1 \in \pop{c'}: \neg \pred{S}(\lv{c'}_1, \ind{N})$ and $\forall \lv{p'}_1 \in \pop{p'}: \neg \pred{S}(\ind{M}, \lv{p'}_1)$ respectively. Since these clauses now contains only one literal, we can apply unit propagation on them. Unit propagation on $\pred{S}(\lv{c'}_1, \ind{N})$ and $\neg \pred{S}(\ind{M}, \lv{p'}_1)$ will give the following theory:
\[\forall \lv{c'} \in \pop{c'} \exists \lv{p'} \in \pop{p'}: \pred{S}(\lv{c}, \lv{p'})\]
\[\forall \lv{p'} \in \pop{p'} \exists \lv{c'} \in \pop{c'}: \pred{S}(\lv{c}, \lv{p'})\]
\[\forall \lv{p'} \in \pop{p'}, \lv{c'}_1, \lv{c'}_2 \in \pop{c'}: \neg \pred{S}(\lv{c'}_1, \lv{p'}) \vee \neg \pred{S}(\lv{c'}_2, \lv{p'})\]
\[\forall \lv{p'}_1, \lv{p'}_2 \in \pop{p'}, \lv{c'} \in \pop{c'}: \neg \pred{S}(\lv{c'}, \lv{p'}_1) \vee \neg \pred{S}(\lv{c'}, \lv{p'}_2)\]
This theory is identical to the original theory that we started with, except that it is over the smaller populations $\pop{c'}$ and $\pop{p'}$. By continuing this process and keeping the intermediate results in a cache, the \wmc\ of the theory can be found in $O(n)$: the number of times we do the above procedure is at most $n=|\pop{c}|$ and each time we do the procedure it takes a constant amount of time. \hfill$\qedsymbol$

\bibliography{MyBib}
\bibliographystyle{aaai}

\end{document}